\documentclass[conference]{IEEEtran}
\IEEEoverridecommandlockouts
\usepackage{cite}
\usepackage{amsmath,amssymb,amsfonts}
\usepackage{algorithmic}
\usepackage{graphicx}
\usepackage{textcomp}
\usepackage{caption}
\usepackage{subcaption}
\usepackage{multirow}
\usepackage{boldline} 
\usepackage{subcaption,mwe}
\usepackage{soul}
\usepackage{tabularx}
\usepackage{booktabs,chemformula}
\usepackage{mathtools}
\usepackage{makecell}
\usepackage{adjustbox}

\usepackage{xcolor}
\def\BibTeX{{\rm B\kern-.05em{\sc i\kern-.025em b}\kern-.08em
    T\kern-.1667em\lower.7ex\hbox{E}\kern-.125emX}}

\begin{document}

\title{Microscale 3-D Capacitance Tomography with a CMOS Sensor Array \vspace*{-5mm}
\thanks{This work was supported by the National Science Foundation under Grant No. 2027108. J.W.L. also acknowledges support from NIGMS 1R35GM142584-01 and the Burroughs Wellcome Fund. We thank P. Joshi for assistance with electron microscopy.}}

\author{\vspace*{-2mm}
Manar Abdelatty$^{1}$, Joseph Incandela$^{2}$, Kangping Hu$^{1}$, Joseph W. Larkin$^{2}$, Sherief Reda$^{1}$, and Jacob K. Rosenstein$^{1}$ \\ \\ $^{1}$Brown University, Providence, RI, USA \\ $^{2}$Boston University, Boston, MA, USA\vspace*{-2mm}}

% \textsuperscript{1} School of [School Name]
% \and
% \and
% \and
% \and
% \and

\maketitle

\begin{abstract}
% This document is a model and instructions for \LaTeX.
% This and the IEEEtran.cls file define the components of your paper [title, text, heads, etc.]. *CRITICAL: Do Not Use Symbols, Special Characters, Footnotes, 
% or Math in Paper Title or Abstract.
Electrical capacitance tomography (ECT) is a non-optical imaging technique in which a map of the interior permittivity of a volume is estimated by making capacitance measurements at its boundary and solving an inverse problem. While previous ECT demonstrations have often been at centimeter scales, ECT is not limited to macroscopic systems. In this paper, we demonstrate ECT imaging of polymer microspheres and bacterial biofilms using a CMOS microelectrode array, achieving spatial resolution of 10 microns. Additionally, we propose a deep learning architecture and an improved multi-objective training scheme for reconstructing out-of-plane permittivity maps from the sensor measurements. Experimental results show that the proposed approach is able to resolve microscopic 3-D structures, achieving 91.5\% prediction accuracy on the microsphere dataset and 82.7\% on the biofilm dataset, including an average of 4.6\% improvement over baseline computational methods.

\end{abstract}


\section{Introduction}
Electrical capacitance tomography (ECT) is an imaging technique that estimates the internal distribution of permittivity in a volume by measuring capacitance between electrodes placed at its boundary~\cite{ect-physics}. It is closely related to electrical impedance tomography (EIT), which estimates the conductivity distribution using impedance measurements~\cite{eit-siam}. Both of these techniques are useful in applications where there is spatial contrast in conductivity or permittivity, including organ and tissue imaging~\cite{Adler-2017-eitapp,rosa2020bladder,Cherepenin_2001_cancer,Warsito_2012_ect_cancer,wu2018human}, neural imaging and neural activity monitoring~\cite{Aristovich_2016_eit_neural_activity,arshad2015towards},~\cite{waristo_2013_ect_neueral_activity}, and
industrial process monitoring of fluid flows~\cite{Li_2013_eit_flow},~\cite{Perera_2017_ect_flow}.  

The physics of the ECT problem, in 2-D, are managed by the Poisson PDE in Eq.~\ref{eq:1}, where $\sigma (x,y)$ represents the permittivity distribution and $u(x,y)$ represents the electric potential~\cite{Alme-2006-ect}. Mutual capacitance $C_{ij}$ between electrodes $i$, $j$ is evaluated by the integral in Eq.~\ref{eq:2}, where $V_{ij}$ is the potential difference between the two electrodes, and $S$ is the path enclosing the sensing electrodes. The problem of estimating the capacitance given the permittivity distribution is referred to as the \emph{forward problem}~\cite{ect-problem}. Conversely, estimating the permittivity distribution from boundary capacitance measurements is referred to as the \emph{inverse problem}. 

\vspace*{-3mm}
\begin{equation} \label{eq:1}
    \nabla . (\sigma(x,y) \nabla u(x, y)) = 0 
    \vspace*{-2mm}
\end{equation}

\begin{equation} \label{eq:2}
    C_{ij} = -\frac{Q}{V_{ij}} = -\frac{\oint_s \sigma(x, y) \nabla u(x,y) ds}{u_i - u_j } 
    \vspace*{-0.5mm}
\end{equation}

The \emph{inverse problem} of ECT is a non-linear and severely ill-posed problem, without unique numerical solutions\cite{Yang_2003_ect_algo}. Therefore, regularization priors are often used to impose an additional constraint on the estimated solution~\cite{ect-reg}. Traditional algorithms solve the inverse problem by minimizing a least squares objective with an additional regularization term for an initial permittivity distribution through an iterative solver ~\cite{Yang_2003_ect_algo,Yang_1999_landweber,Thusara_2012_TV}. However, iterative algorithms are sensitive to noise in the capacitance measurements which makes them more susceptible to divergence. Prior work demonstrates that deep learning models can be more robust to experimental noise and can provide accurate image reconstructions~\cite{CNN-AE},~\cite{Zheng},~\cite{Zhu},~\cite{transformer}. 

% \begin{figure}[!b]
%     \centering
%     \includegraphics[width=\linewidth]{figures/exp_setup/concept.pdf} 
%     \captionsetup{font={small}}
%     \caption{Electrical capacitance tomography (ECT) uses capacitance measurements at a boundary to estimate the permittivity map within a sample.}
%     \label{fig:concept}
% \end{figure}

Previous demonstrations of ECT have often resolved centimeter-scale targets. If it could be appropriately miniaturized, one appealing application of ECT would be for 3-D visualization of cell cultures~\cite{linderholm2007cell,yang2016miniature,xian2023onChipECT}. Optical confocal microscopy is a powerful tool for biologists to image the 3-D structure of complex cell cultures \cite{schlafer2017confocal}. However, confocal imaging can be prohibitively expensive for routine use, usually relies on fluorescent labeling, and its intense light excitation introduces tradeoffs between the frame rate and risks of photobleaching and phototoxicity.

Here we propose a microscale capacitance tomography system using a 512\,$\times$\,256 CMOS sensor array~\cite{hu2022cicc,cmos-sensor-tbiocas,hu2021super}, achieving the highest-resolution ECT reported to date (10\,$\mu$m). We apply deep learning to approximate the ECT inverse operator, using training data as a regularization prior to the ill-posed inverse problem. The proposed system enables imaging of micro 3-D structures of cell cultures with a high reconstruction accuracy. We present results for two experimental datasets of microscopic objects: polymer microspheres and bacterial biofilms. The ECT data are trained and evaluated using ground truth images acquired from 3-D confocal microscopy.

\begin{figure*}[!h]
\vspace*{-2mm}
  \includegraphics[width=\textwidth]{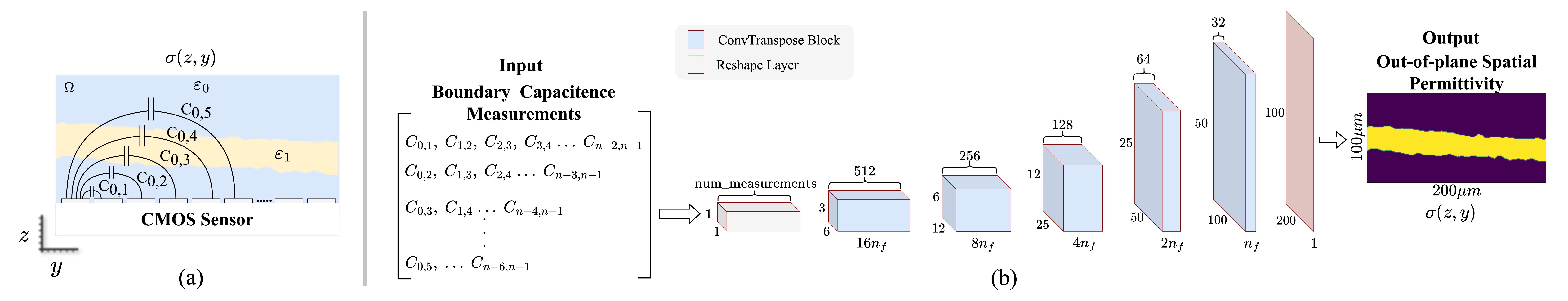}
  \captionsetup{font={small}}
  % \vspace*{-6mm}
\captionsetup{font={small}}
\vspace{-6mm}
\caption{Overview of the tomography system. (a) Illustration of the one-sided planar ECT detection using the CMOS sensor. The sample, above the sensor, has a permittivity value $\varepsilon_1$ distinct from the background permittivity $\varepsilon_0$. (b) Image reconstruction network, based on transposed convolution. The input is a matrix of pairwise capacitance measurements acquired from the CMOS sensor. The output is a $100$\,$\times$\,$200\,\mu m$ cross-sectional image that represents the permittivity distribution $\sigma (z,y)$.}\label{fig:1}  
\vspace*{-4mm}
\end{figure*}

% \section{Hardware Platform Design}

% The tomography is implemented using measurements from an integrated CMOS microelectrode array described in \cite{hu2022cicc}. This chip has a 512\,$\times$\,256 planar array of microelectrodes on a 10\,$\mu$m\,$\times$\,10\,$\mu$m rectangular grid. In one of its operating modes, the chip can efficiently measure the mutual capacitance between any two pixels in the array \cite{cmos-sensor-tbiocas}. An image of the sensor is shown in Fig.~\ref{fig:exp_setup}a.
% %\JKR{more here, and a figure}

% % More here 
% % Explain more about the problem 
% % Reference things in order 
% % 

\section{Methodology}

\subsection{Capacitance Tomography Hardware}
The tomography is implemented using measurements from an integrated CMOS microelectrode array described in \cite{hu2022cicc}. This chip has a 512\,$\times$\,256 planar array of microelectrodes on a 10\,$\mu$m\,$\times$\,10\,$\mu$m rectangular grid. In one of its operating modes, the chip can efficiently measure the mutual capacitance between any two pixels in the array \cite{cmos-sensor-tbiocas,hu2021super}. Fig.~\ref{fig:1}(a) illustrates the one-sided planar detection using the sensor, where electrodes are only placed at the bottom boundary. Each capacitance $C_{ij}$ represents fringing electric fields through the sample, and thus the permittivity and geometry of materials near the sensor influence these measurements. Samples to be imaged are placed on the chip surface in a liquid or gel electrolyte, as illustrated in Fig.~\ref{fig:exp_setup}(a). An image of the sensor is shown in Fig.~\ref{fig:exp_setup}(b).

\subsection{Image Reconstruction Network}

Fig.~\ref{fig:1}(b) illustrates the architecture of the image reconstruction network. The input is an $m \times n$ matrix containing pairwise capacitance measurements, where $m$ is the number of spatial offsets considered when measuring the mutual capacitance values and $n$ is the number of electrodes. Each entry in the matrix corresponds to the mutual capacitance $C_{ij}$ between electrodes $i$ and $j$. %The row indicates the spatial offset at which the capacitance measurements are made. 
For example, the first row contains $n$ capacitance values measured between adjacent electrodes, and the second row contains measurements between electrodes separated by $2$ positions. In this study, we only use capacitance measurements with $|i-j| \le 5$. To make the input matrix compatible with the transposed convolution layer, we reshape it to a 3-D feature map of size $(w,h,c)=(1,1,\textrm{num\_measurements})$. The input 3-D feature map is then repeatedly up-sampled by a factor of $2$ until it reaches the spatial resolution of the predicted cross-sectional image $(w,h,c)=(200,100,1)$, which represents the permittivity distribution $\sigma(z,y)$ of the medium above the CMOS sensor.  

% \JKR{approximately? For odd numbers like 12 to 25 the upsampling factor is slightly larger than 2, but I guess we could just ignore that because the factor is actually 2 and I just zero-pad to change the output dimension from 24 to 25}

The boundary capacitance measurements are up-sampled through a series of five transposed convolution blocks. Each block contains a transposed convolution layer, batch normalization layer, and a ReLU activation, except for the last block where sigmoid activation is used to constrain the output permittivity to be in the range $[0, 1]$. The transposed convolution layer contains a learnable kernel that is used to reconstruct a high-resolution output from a low-resolution input~\cite{dumoulin2016guide}. Batch normalization is used for training stability and convergence speed-up.
%~\cite{sergey-2015-batchnorm}. 
Additionally, a residual connection is added between the block input and output through a $1\text{x}1$ convolution kernel.

\subsection{Loss}
% https://arxiv.org/pdf/2102.04525v4.pdf
The loss function is important in training deep learning algorithms as it defines the optimization landscape and has a significant impact on the model convergence~\cite{Yeung_2022_unified_focal_loss}. Class imbalance, where the foreground permittivity occupies a significantly smaller region relative to the background pixels, poses a challenge in training. As noted by~\cite{Zhu}, distribution-based loss functions like the focal loss~\cite{focal-loss} can help address the class-imbalance issue. However, region-based losses and compound losses have been shown to consistently provide better performance than distribution-based losses~\cite{Yeung_2022_unified_focal_loss}. Therefore, we propose a compound loss function, shown in Eq.~\ref{loss:4}, that combines a distribution-based loss (Focal Loss  $L_{\text{FL}}$), a region-based loss (Dice Loss $L_{\text{Dice}}$~\cite{dice-loss}), and a pixel-to-pixel loss (Smooth L1 Loss $L_{\text{SmoothL1}}$~\cite{Ross_2015_fast_rcnn}). The weighting parameters ($\lambda_1, \lambda_2, \lambda_3$) define the tradeoff between the different loss-objectives and are learned during training.

% The model is trained with a multi-objective loss function. Each objective works on a different scale to improve the reconstruction quality of the predicted images. The compound loss function, shown in Eq.~\ref{loss:4}, combines a distribution-based loss function (Focal Loss $L_{FL}$~\cite{focal-loss}), a region-based loss function (Dice Loss $L_{Dice}$~\cite{dice-loss}), and a pixel-to-pixel loss function (Smooth L1 Loss $L_{SmoothL1}$~\cite{Zhu}), where $y$ is the ground truth cross-sectional image and $\hat{y}$ is the predicted cross-sectional image. The weighting parameters ($\lambda_1, \lambda_2, \lambda_3$) are learned during training. 

\vspace*{-5mm}
\begin{equation}\label{loss:4}
\resizebox{.91\hsize}{!}{$L(y,\hat{y}) = \lambda_1 L_{\text{SmoothL1}} (y, \hat{y}) + \lambda_2 L_{\text{FL}}(y, \hat{y}) + \lambda_3 L_{\text{Dice}}(y, \hat{y}) $}
\end{equation}
\vspace*{-5.1mm}

% \vspace*{-4mm}
% \begin{equation}\label{loss:4}
% {\fontsize{9}{12}\selectfont
% \begin{aligned}
% L(y,\hat{y}) &= \lambda_1 L_{\text{SmoothL1}} (y, \hat{y}) + \lambda_2 L_{\text{FL}}(y, \hat{y}) + \lambda_3 L_{\text{Dice}}(y, \hat{y})
% \end{aligned}
% \tag*{\small{(4)}}
% }
% \end{equation}

Each loss component in Eq.~\ref{loss:4} represents a different objective that the model aims to optimize. The smooth L1 loss measures the absolute difference between the predicted and ground truth images, with added smoothing to make it differentiable and less sensitive to outliers. It addresses pixel-level differences and equally penalizes the error in the foreground and background pixels. The focal loss is a modified cross-entropy loss that dynamically scales during training to help the model focus on the hard-to-predict examples. This is done by adding a scaling factor that decays to zero as the model confidence increases in the easy-to-predict examples. Dice loss is used to emphasize the spatial agreement and boundary delineation between the predicted and ground truth images by maximizing the overlap region between the two images.

% https://www.sfu.ca/~abentaie/papers/uncertainty.pdf

\section{Experimental Results}

\begin{figure}[htbp]
\vspace*{-2mm}
\centering
\begin{subfigure}[b]{0.45\linewidth}
    \centering
    \begin{subfigure}[b]{\linewidth}
     \includegraphics[width=\linewidth]{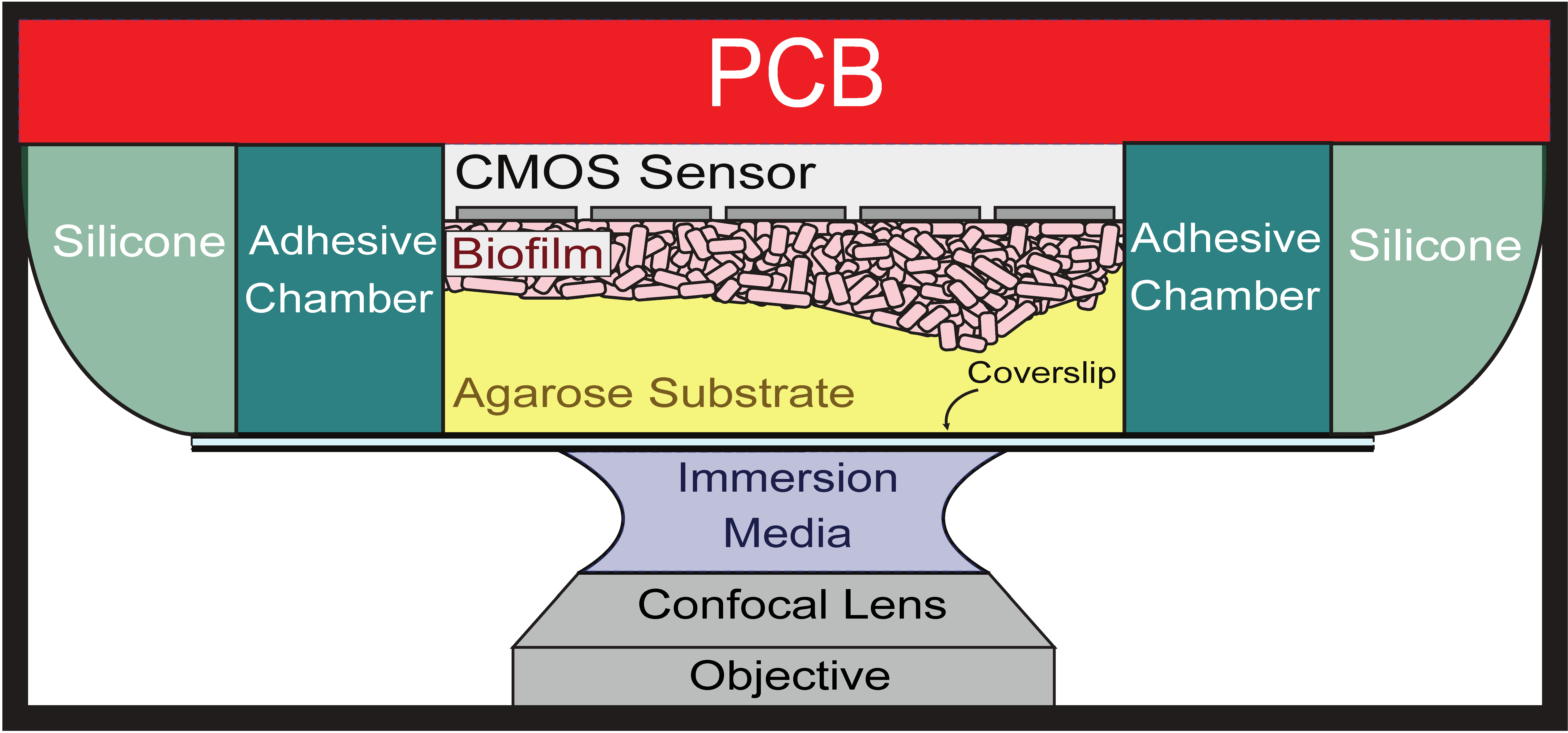}
      \captionsetup{font={footnotesize}}
    \vspace*{-5.5mm}
      \caption{}
    \label{fig:plot1}
    \end{subfigure}
    % Subplot 1
    \vspace{0.cm} % Adjust vertical spacing between subplots
    \begin{subfigure}[b]{\linewidth}
    \includegraphics[width=\linewidth]{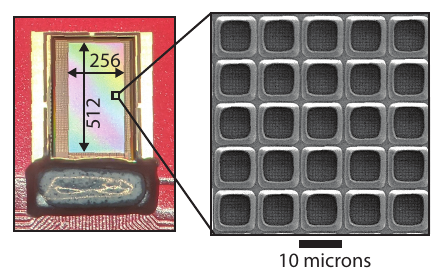} % Subplot 2
    \vspace*{-7.8mm}
   \captionsetup{font={footnotesize}}
    \caption{}
    \end{subfigure}
\end{subfigure}%
\begin{subfigure}[b]{0.274\linewidth}
    \begin{subfigure}[b]{\linewidth}
        \centering
        \includegraphics[width=\linewidth]{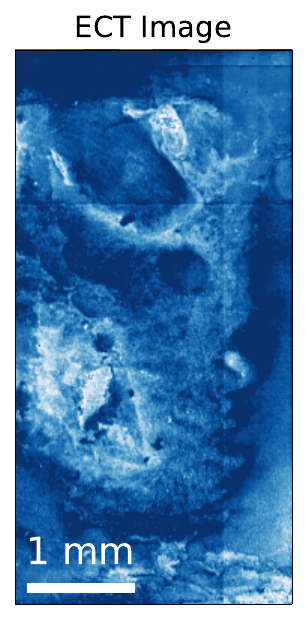} % Subplot 3
  \vspace*{-5.4mm}
  \captionsetup{font={footnotesize}}
   \caption*{\hspace{2.2 cm}  (c)}
    \end{subfigure}
\end{subfigure}%
\begin{subfigure}[b]{0.274\linewidth}
    \begin{subfigure}[b]{\linewidth}
    \centering
    \includegraphics[width=\linewidth]{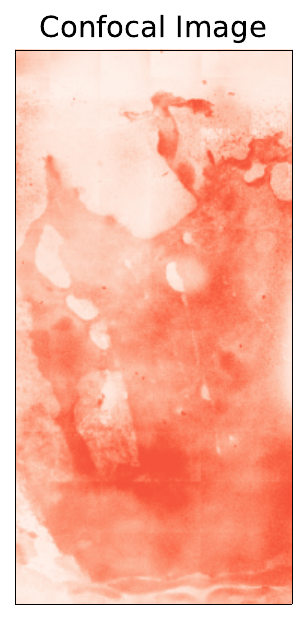} % Subplot 4
  \vspace*{-5.4mm}
  \captionsetup{font={footnotesize}}
    \caption*{}
    \end{subfigure}
\end{subfigure}
  \captionsetup{font={small}}
  \vspace{-6mm}
  \caption{(a) Experimental samples were mounted to allow both ECT measurements as well as 3-D optical images from a confocal microscope. (b) The CMOS microelectrode array has 131,072 electrodes on a 10 micron grid. (c) A dataset with a \emph{B. subtilis} biofilm showing the confocal max projection (right) and one mutual capacitance image measured using a spatial offset of $1$ (left).}
  \label{fig:exp_setup}
  \vspace*{-3.4mm}
\end{figure}

In order to obtain both ECT data and confocal 3-D images of the same objects, we sealed samples on the sensor with optically transparent windows, as shown in Fig.~\ref{fig:exp_setup}(a). The confocal images are useful as a ground truth for training the inverse algorithms, as well as for qualitative and quantitative comparisons of the reconstructed sample geometry. Using this setup, we performed experiments with both polymer micropheres and bacterial biofilms.

\subsection{Polymer Microspheres}

A sample containing 30\,$\mu$m purple fluorescent polystyrene microspheres (Spherotech Inc., IL, USA) was positioned over the CMOS array. As shown in Fig.~\ref{fig:exp_setup}(a),
a 500\,$\mu$m deep well was created around the CMOS sensor with a stack of two 25\,$\mu$L adhesive chambers (Gene Frame, Thermo Scientific). Microspheres were added to a buffered Minimal Salts Glycerol Glutamate (MSGG) media in a 10$\times$ dilution, along with agarose flakes. The mixture was autoclaved and 50\,$\mu$L of the hot solution was pipetted into the well, covering the sensor. The well was then sealed with a 22\,mm\,$\times$\,22\,mm coverslip, and the solution was allowed to solidify into a 2\% agarose gel which immobilized the microspheres. Finally, the edges of the assembly were sealed with a fast-setting silicone elastomer (EcoFlex 5, Smooth-On, Inc.) to prevent the gel from drying which would introduce distortions during the imaging process.% A cross-section of the assembly is illustrated in Fig.~\ref{fig:exp_setup}(a).

\begin{figure}[!h]
\vspace*{-2mm}
\centering

\begin{subfigure}[t]{\linewidth}
\centering
    \begin{subfigure}[t]{.4\linewidth}
    \centering
    \includegraphics[width=\linewidth]{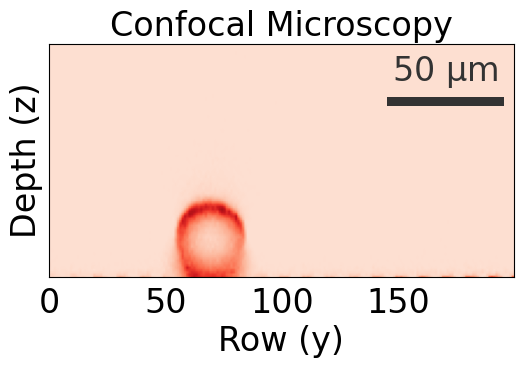}
    \vspace{-6.8mm}
  \captionsetup{font={footnotesize}}
    \caption*{\hspace{3.45 cm}  (a)}
    \vspace{-10mm}
    \label{fig:cost_breakdown} 
\end{subfigure}
%\begin{subfigure}[t]{.325\linewidth}
%   \centering
%   \includegraphics[width=\linewidth, height=1.8cm]{figures/beads/1_cross_section_1310.png}
%    \captionsetup{font={small}}
%    \vspace*{-6mm}
%    \caption{}
%   \label{fig:cost_breakdown_forecast}
%\end{subfigure}
\begin{subfigure}[t]{.4\linewidth}
   \centering
  \includegraphics[width=\linewidth]{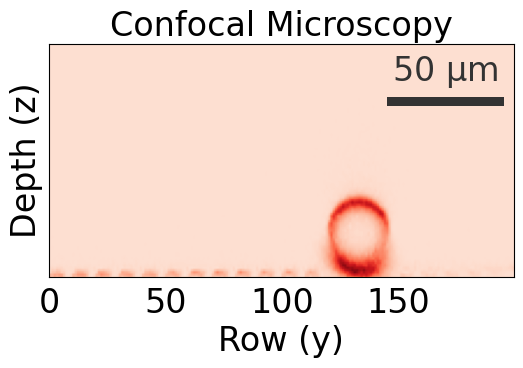}
   % \caption{FNN-AE}
   \label{fig:cost_breakdown_forecast}
\end{subfigure}
\end{subfigure}

\begin{subfigure}[t]{.4\linewidth}
   \centering
   \includegraphics[width=\linewidth]
   {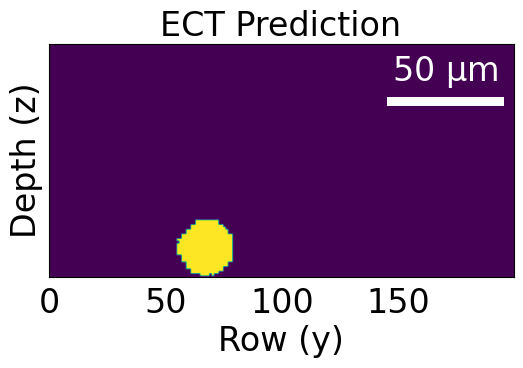}
   \label{fig:cost_breakdown_forecast}
  \captionsetup{font={footnotesize}}
   \vspace{-6.8mm}
    \caption*{\hspace{3.45 cm}  (b)} 
\end{subfigure}
%\begin{subfigure}[t]{.325\linewidth}
%   \centering
%   \includegraphics[width=\linewidth, height=1.8cm]
%   {figures/beads/pred_2_0.png}
%   \captionsetup{font={small}}
%   \vspace*{-6mm}
%  \captionsetup{font={small}}
%   \caption{}
%   \label{fig:cost_breakdown_forecast}
%\end{subfigure}
\begin{subfigure}[t]{.4\linewidth}
   \centering
   \includegraphics[width=\linewidth]
   {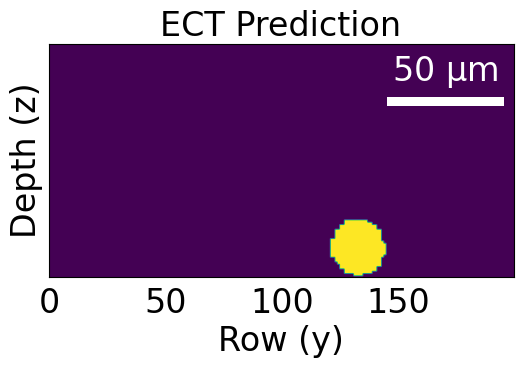}
   \label{fig:cost_breakdown_forecast}
\end{subfigure}
\captionsetup{font={small}}
\vspace{-3mm}
\caption{Image reconstruction of polymer microspheres (a) Confocal microscopy ground truth. (b) ECT model prediction.}
\label{fig:bead_plot}
\vspace*{-4mm}
\end{figure}

Due to the sparse distribution of the polymer microsphere on the chip surface, we obtained a limited number of capacitance and confocal cross-sectional image pairs (16 different pairs). Therefore, we augmented our dataset with a synthetic dataset of 5,975 capacitance and cross-sectional image pairs generated using pyEIT~\cite{pyeit}, which runs finite element electrostatic simulations that solve the PDE in Eq.~\ref{eq:1}. The synthetic dataset was mainly used for training, while the experimental dataset was used for testing. In order to make the model more robust to the noise present in the experimental data, the simulated capacitance values were perturbed with a Gaussian noise $\epsilon_i\in \mathcal{N}(0, 0.03)$ during training. Fig.~\ref{fig:bead_plot} shows the model predictions on the experimental microsphere testing dataset. The results demonstrate the system's ability to accurately predict the shape and location of the microsphere from the experimental ECT measurements, despite being trained solely on synthetic data.     

% The results indicate that the system can capture the correct shape and location of the bead from the experimental ECT measurement, even though it is only trained with synthetic data.    

\newcommand\Tstrut{\rule{0pt}{2.2ex}}         % = `top' strut
\newcommand\Bstrut{\rule[-1.2ex]{0pt}{0pt}}   % = `bottom' strut

\begin{table*}[ht]
\centering
\captionsetup{font={small}}
\vspace*{-2mm}
\caption{\label{ablation} Comparison to prior electrical capacitance tomography (ECT) systems. }
\vspace{-1mm}
\begin{adjustbox}{max width=1\textwidth,center}
\begin{tabular}{|c|c|c|c|c|c|c|}
\hlineB{}
& \cite{Zheng} \Tstrut\Bstrut & \cite{Zhu} & \cite{transformer} & \cite{Suo}   & \cite{xian2023onChipECT} & \textbf{This Work} \\
\hline
Imaging Domain \Tstrut\Bstrut & Circular & Circular  & Circular & Planar & Planar & Planar\\ [0.5ex]
Reconstruction Algorithm \Bstrut& FNN-AE & FNN+CNN-AE & self-attn+UNet & Tikhonov & Linear Back-projection& TCNN+MOL\\[0.5ex]
Imaging Application \Bstrut & 3D Objects & 3D objects & Cryogenic Fluids  & 3D objects & Yeast cells & Bacterial biofilms\\[0.5ex]
Electrode Size \Bstrut & $cm$ scale & $cm$ scale & $cm$ scale & \makecell{$mm$  scale (6x6 $mm^2$)} & \makecell{$mm$ scale (1.4x0.8$mm^2$)}   & \makecell{$\mu m$ scale  (10x10$\mu m^2$)}  \\[0.5ex] 
Array Size & 8 electrodes  &  16 electrodes &  8 electrodes & 16 electrodes (4$\times$4) & 34 electrodes & 131,072 electrodes (512$\times$256) \\[0.5ex]
\hline
\end{tabular}
\end{adjustbox}
\vspace{-2mm}
\end{table*}

\subsection{Bacterial Biofilm}

To further develop the tomography capabilities, we aimed to produce 3-D maps of biomass within bacterial biofilms. \textit{Bacillus subtilis} biofilms were grown on 500 $\mu$m-thick substrates of 1.8\% agarose MSGG media for roughly 24 hours. A biofilm was cut out along with a thin supporting agarose pad and transferred onto the CMOS sensor. Before transferring the biofilm, the sensor was treated with poly-L-lysine to improve cell adhesion. The biofilm was sealed with a glass coverslip and fast-setting silicone elastomer to prevent drying, and mounted on a confocal microscope (Stellaris 5, Leica). An illustration of the completed device is shown in Fig.~\ref{fig:exp_setup}(a). 

\begin{figure}[!b]
\centering
\vspace{-3mm}
\begin{subfigure}[t]{.325\linewidth}
    \centering
    \includegraphics[width=\linewidth]{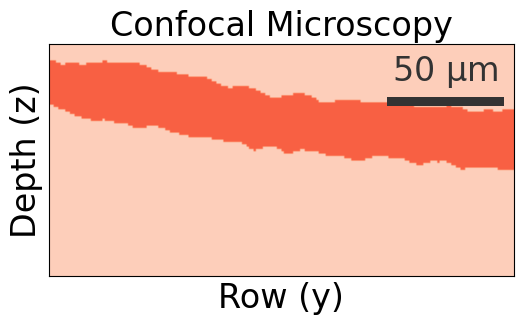}

    % \caption{Ground Turth}
    \label{fig:cost_breakdown} 
\end{subfigure}
\begin{subfigure}[t]{.325\linewidth}
   \centering
   \includegraphics[width=\linewidth]{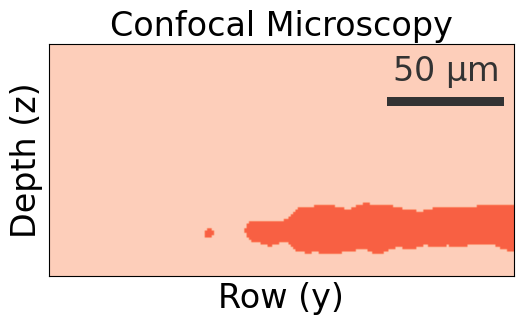}
   \captionsetup{font={footnotesize}}
   \vspace*{-6mm}
  \captionsetup{font={footnotesize}}
   \caption{}
   \label{fig:cost_breakdown_forecast}
\end{subfigure}
\begin{subfigure}[t]{.325\linewidth}
   \centering
  \includegraphics[width=\linewidth]{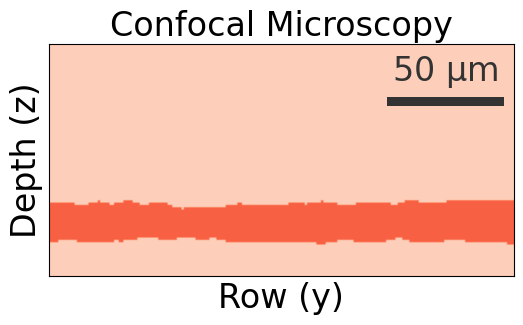}
   % \caption{FNN-AE}
   \label{fig:cost_breakdown_forecast}
\end{subfigure}

\begin{subfigure}[t]{.325\linewidth}
   \centering
\includegraphics[width=\linewidth]
   {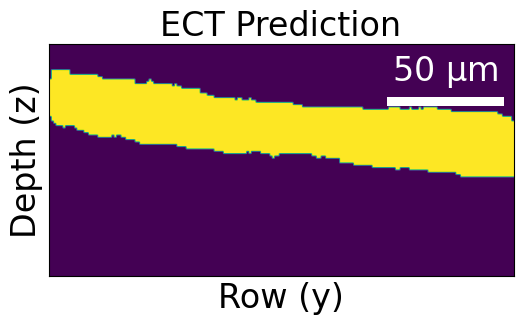}
   \label{fig:cost_breakdown_forecast}
\end{subfigure}
\begin{subfigure}[t]{.325\linewidth}
   \centering
   \includegraphics[width=\linewidth]
   {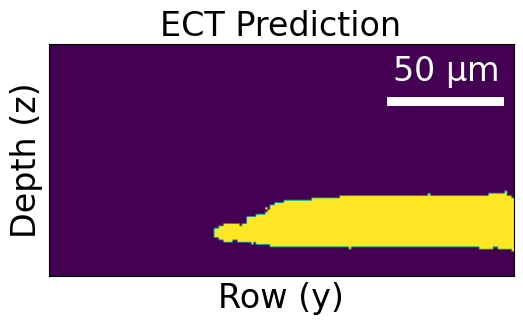}
   \captionsetup{font={footnotesize}}
   \vspace*{-6mm}
  \captionsetup{font={footnotesize}}
   \caption{}
   \label{fig:cost_breakdown_forecast}
\end{subfigure}
\begin{subfigure}[t]{.325\linewidth}
   \centering
    \includegraphics[width=\linewidth, height=1.9cm]
   {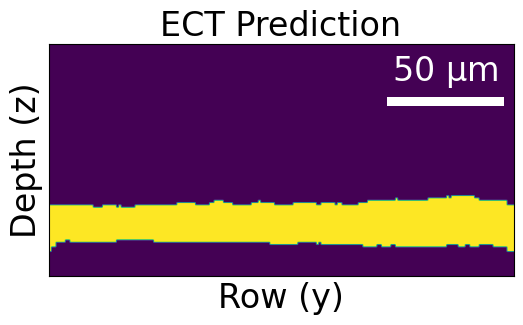}
   \label{fig:cost_breakdown_forecast}
\end{subfigure}
  \captionsetup{font={small}}

\vspace{-2mm}
\caption{Reconstructed permittivity images of sections of a bacterial biofilm (a) Confocal ground truth. (b) ECT model prediction.}
 \label{fig:biofilm_plot}
 \vspace{-2mm}
\end{figure}

From the experimental biofilm dataset shown in Fig.~\ref{fig:exp_setup}(b), we generated 6,400 capacitance and confocal cross-sectional image pairs, which were divided into 80\% training, 10\% validation, and 10\% testing. Fig.~\ref{fig:biofilm_plot} shows the model predictions on the testing set. The results indicate that the system can accurately predict the biofilm thickness, shape, and depth from the experimental ECT measurements. 

\begin{figure}[!h]
\centering
\vspace*{-1.mm}
\begin{subfigure}[t]{\linewidth}
    \centering
    \includegraphics[width=\linewidth]{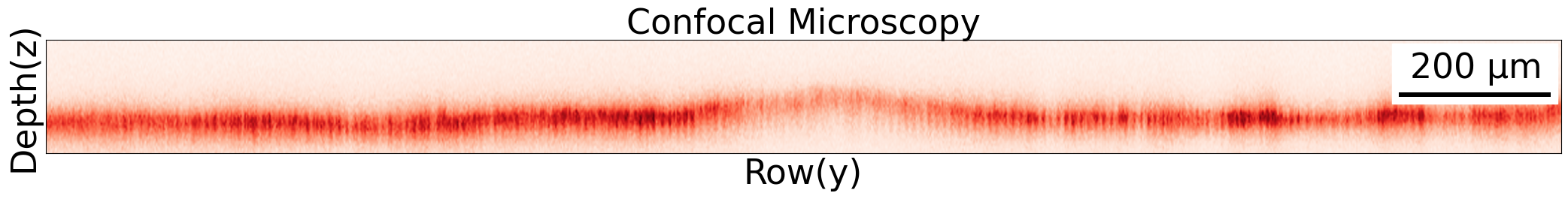}
    \label{fig:cost_breakdown} 
    \vspace*{-6mm}
    \captionsetup{font={footnotesize}}
    \caption{}
\end{subfigure}
\begin{subfigure}[t]{\linewidth}
   \centering
   \includegraphics[width=\linewidth]
    {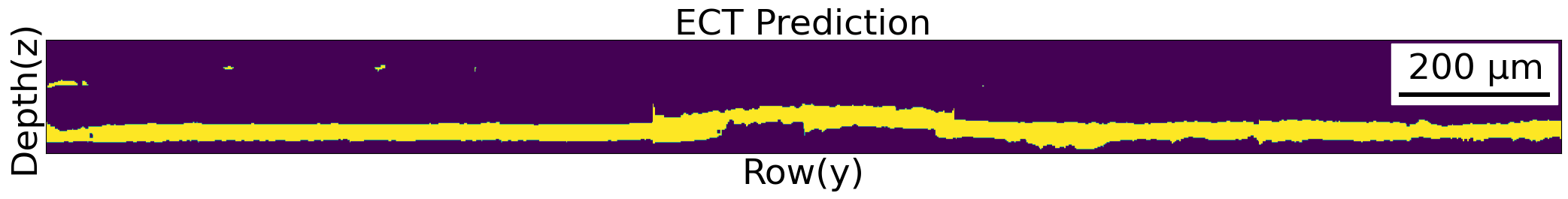}
   \label{fig:cost_breakdown_forecast}
    \vspace*{-6mm}
    \captionsetup{font={footnotesize}}
    \caption{}
\end{subfigure}
\vspace*{-1mm}
\captionsetup{font={small}}
\vspace{-1mm}
\caption{Reconstruction of a larger-scale cross-sectional image of a \emph{B. subtilis} biofilm spanning 2\,mm. (a) Confocal ground truth. (b) Model prediction, stitched together from ten 200\,$\mu$m windows.}
% \vspace*{-0.5mm}

\label{fig:column_plot}
\vspace{-7mm}
\end{figure}

Predictions were performed independently on 100\,$\mu$m\,$\times$\,200\,$\mu$m  meshes. However, we can reconstruct larger areas by simply stitching the predicted local meshes together. Fig.~\ref{fig:column_plot} shows the model predictions along a linear array of 200 electrodes (2\,mm total length), demonstrating the system's ability to resolve larger millimeter-scale features in the biofilm.  

\subsection{Baselines}
\begin{table}[!b]
  \centering
\captionsetup{font={small}}
\vspace{-2mm}
\caption{\label{sota-table} Quantitative comparison to prior work using the microsphere ($\Omega_1$) and biofilm ($\Omega_2$) datasets.}
\vspace{-1mm}

  \begin{tabular}{@{}ccccccc@{}}
    \toprule
    & Dataset & MSE $\downarrow$  & SSIM $\uparrow$ & PSNR $\uparrow$ & CC $\uparrow$ & IoU $\uparrow$  \\
    \midrule
    % Tikhonov & Metallic & 0.00 \\
    \multirow{2}{1.4cm}{Tikhonov} 
& $\Omega_1$ & 0.315 & 0.040 & 5.014 & 0.207 & 0.485\\ 
& $\Omega_2$ & 0.115 & 0.569 & 9.395 & 0.171 & 0.362   \\
    \midrule
\multirow{2}{1.3cm}{FNN-AE ~\cite{Zheng}} 
& $\Omega_1$ & 0.026 & 0.422 & 15.732 & 0.238 & 0.485 \\ 
& $\Omega_2$ & 0.145 & 0.678 & 8.360 & 0.447 &  0.591  \\
    \midrule
\multirow{2}{1.4cm}{FNN+CNN-AE~\cite{Zhu}} 
& $\Omega_1$ & 0.017  & 0.930 & 17.452 & 0.679 & 0.765  \\ 
& $\Omega_2$ & 0.124 & 0.656 & 9.043 & 0.615 &  0.685   \\
    \midrule
\multirow{2}{1.4cm}{self-attn+UNet~\cite{transformer}} 
& $\Omega_1$ & 0.005 & 0.914 & 22.681 & 0.898 & 0.679  \\ 
& $\Omega_2$ & 0.071 & 0.784 & 11.478 & 0.694 & 0.775   \\ 
    \midrule

\multirow{2}{1.4cm}{TCNN+MOL (Ours)} 
& $\Omega_1$  & \textbf{0.004} & \textbf{0.975} & \textbf{23.036} & \textbf{0.910} &  \textbf{0.915} \\ 
& $\Omega_2$ & \textbf{0.056}  & \textbf{0.799} & \textbf{12.473} & \textbf{0.781} & \textbf{0.827}  \\ 

    \bottomrule
  \end{tabular}
\vspace*{-0.2mm}
\end{table}

% We compare the reconstructions of the proposed model with one of the traditional algorithms:  iterative Tikhonov and three deep learning based models: a fully-connected auto-encoder (FNN-AE)~\cite{Zheng}, the permittivity value prediction network presented in ~\cite{Zhu} which is composed of two fully connected networks and a post-processing convolutional-based auto-encoder (FNN+CNN-AE), and the UNet architecture presented in~\cite{transformer}. Quantitative results are shown in Table.~\ref{sota-table} and qualitative comparisons are displayed in Fig.~\ref{fig:sota_plot}. 

\begin{figure}[!t]
\centering
\vspace*{-2mm}
\begin{subfigure}[t]{.325\linewidth}
    \centering
    \includegraphics[width=\linewidth]{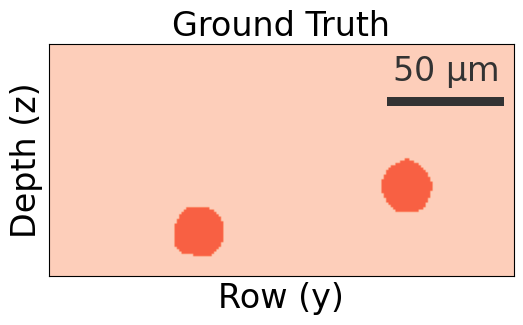}
    \centering
   \captionsetup{font={footnotesize}}
    \vspace{-5.5mm}
    \caption{Ground Truth}
    \vspace*{1.5mm}
    \label{fig:cost_breakdown} 
\end{subfigure}
\begin{subfigure}[t]{.325\linewidth}
   \centering
   \includegraphics[width=\linewidth]{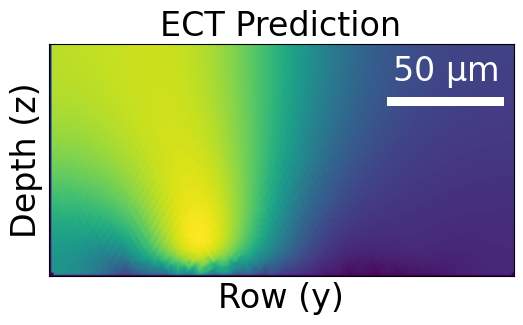}
  \captionsetup{font={footnotesize}}
    \vspace{-5.5mm}
   \caption{Tikhonov}
   \label{fig:cost_breakdown_forecast}
\end{subfigure}
\begin{subfigure}[t]{.325\linewidth}
   \centering
   \includegraphics[width=\linewidth]{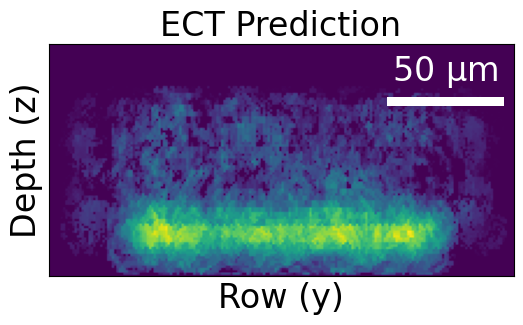}
  \captionsetup{font={footnotesize}}
    \vspace{-5.5mm}
   \caption{FNN-AE}
   \label{fig:cost_breakdown_forecast}
\end{subfigure}

\begin{subfigure}[t]{.325\linewidth}
   \centering
   \includegraphics[width=\linewidth]{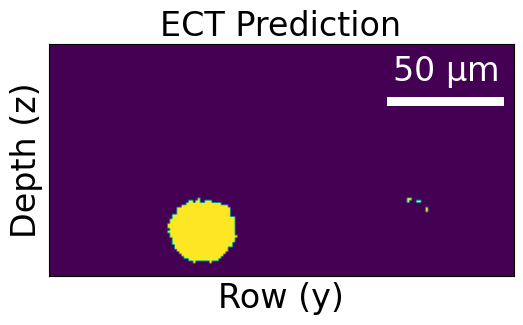}
   \captionsetup{font={footnotesize}}
    \vspace{-5.5mm}
   \caption{FNN+CNN-AE}
   \label{fig:cost_breakdown_forecast}
\end{subfigure}
\begin{subfigure}[t]{.325\linewidth}
   \centering
   \includegraphics[width=\linewidth]{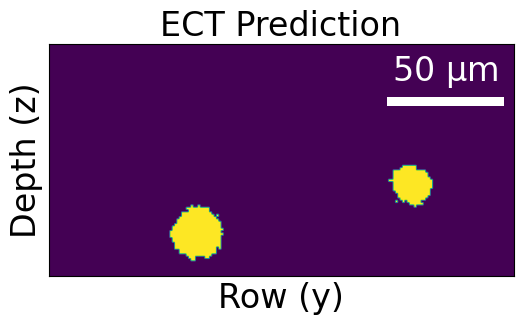}
  \captionsetup{font={footnotesize}}
    \vspace{-5.5mm}
   \caption{self-attn+UNet}
   \label{fig:cost_breakdown_forecast}
\end{subfigure}
\begin{subfigure}[t]{.325\linewidth}
   \centering
   \includegraphics[width=\linewidth]{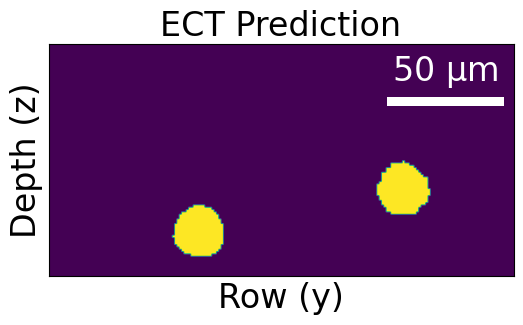}
  \captionsetup{font={footnotesize},width=\linewidth}
    \vspace{-5.5mm}
   \caption{ TCNN+MOL (Ours)}
   \label{fig:cost_breakdown_forecast}
\end{subfigure}
  \captionsetup{font={small}}
\caption{Qualitative comparison to prior algorithms, using a scene of two microspheres simulated using pyEIT \cite{pyeit}. }
\vspace*{-2mm}
\label{fig:sota_plot}
\vspace*{0mm}
\end{figure}

We compare the reconstructions of the proposed model on the microsphere and biofilm testing datasets with one more traditional algorithm (iterative Tikhonov) and three deep learning algorithms which include the fully-connected auto-encoder (FNN-AE) presented in~\cite{Zheng}, the permittivity prediction network presented in~\cite{Zhu} which is composed of two fully connected networks and a post-processing convolutional-based auto-encoder (FNN+CNN-AE), and the self-attention and UNet-based model (self-attn+UNet) presented in~\cite{transformer}. Quantitative comparisons are performed using mean squared error (MSE), and a set of perceptual metrics including structural similarity index measure (SSIM), peak signal-to-noise-ratio (PSNR), cross-correlation (CC), and intersection over union (IoU). Quantitative results are shown in Table.~\ref{sota-table} and qualitative comparisons are displayed in Fig.~\ref{fig:sota_plot}.
Judged by the IoU, the overall accuracy is 91.5\% for the microsphere dataset, and 82.7\% for the biofilm dataset.

%~\cite{wang_2004_ssim}
%~\cite{Alain_2010_PSNR}
%~\cite{Hamid_2019_IoU}
%~\cite{Zhao_2006_CC}

The Tikhonov algorithm provides a good estimate for the location of the shallow bead, but fails to recognize sharp boundaries and to predict the deeper bead. This is because for planar electrodes, changes at the boundary are very subtle for relatively deep objects~\cite{Chen_2019_EIT_Depth}, and the Tikhonov algorithm converges to a sub-optimal reconstruction. The FNN-AE falls into a local minimum mainly because fully connected layers are not suitable for the task. While the predictions are improved by the post-processing CNN-AE, the FNN+CNN-AE also fails to predict the deeper bead. The self-attn+UNET can correctly capture the presence of the two beads, however, it underestimates the diameter of the deep bead. This is because the self-attn+UNet model is trained with the MSE loss, which is known to produce blurred/smeared predictions~\cite{Michel_2016_ICLR}. By incorporating a region-based loss that enhances the spatial alignment between the predictions and the ground truth images and a distribution-based loss that addresses the class-imbalance problem, our proposed model (TCNN+MOL) can predict the shape and location of both the shallow and deeper beads
%The improvements in results are mainly attributed to the choice of the loss function
\cite{Taghanaki_2019_combo_loss}. 

\subsection{Ablation Study}

To analyze the effectiveness of the proposed approach, we conduct an ablation study on training the model with different combinations of the loss objective (Table~\ref{loss-ablation}). In this experiment, the model was trained for 20 epochs on the biofilm dataset. We see the lowest CC when the model is trained with a per-pixel loss function ($L_{\text{Smooth L1}}$). CC is improved by adding the focal loss as it helps address the class imbalance issue. The dice loss further improves performance by maximizing overlap between the predictions and the ground truth.  

\begin{table}[h]
\vspace*{-2mm}
\centering
\captionsetup{font={small}}
 \caption{\label{loss-ablation} Ablation on the loss function}
 \vspace{-1mm}
\begin{tabular}{l c} 
 \toprule
 Loss  & Cross Correlation (CC)  \\
 [0.5ex] 
 \midrule
 $L_{\text{Smooth L1}}$ & 0.7363  \\ 
 % \hline
 $L_{\text{Smooth L1}} + L_{\text{FL}}$  & 0.75104 \\
 % \hline
$L_{\text{Smooth L1}} + L_{\text{FL}} + L_{\text{Dice}}$ & 0.76320 \\
 % \hline

 \bottomrule
\end{tabular}
\vspace*{-3.4mm}
\end{table}

\section{Conclusion}
We have presented a microscale electrical capacitance tomography (ECT) system using a CMOS biosensor that can predict the 3-D structure of objects over a large field of view. We proposed a deep learning architecture and a multi-objective training scheme for reconstructing out-of-plane images from the sensor array data. We demonstrated the effectiveness of the proposed approach by imaging polymer microspheres and bacterial biofilms. 
%The proposed framework obtains high quality reconstructions on all metrics evaluated (Table ~\ref{sota-table}). 
Compared to prior demonstrations (Table~\ref{ablation}), this work uses significantly smaller electrodes and achieves finer spatial resolution. Microscale ECT can be applied to a wide range of biomedical applications including low-cost non-optical label-free 3-D monitoring of cell cultures.

{\small
\bibliographystyle{plain}
\bibliography{ref}
}

% \clearpage

% \section{Appendix}
% \textbf{Will be commented out before submission: just keeping it for reference}
% \subsection{Ablation Study}

% To analyze the effectiveness of the proposed approach, we conduct an ablation study on training the model with different combinations of the loss objective. In this experiment, the model is trained for 20 epochs on the biofilm dataset with different combinations of the loss objectives shown. As shown in table \ref{loss-ablation}, we get the lowest CC when we train only with a per-pixel loss function ($L_{\text{Smooth L1}}$). CC is improved by adding the focal loss to the training objective as it helps the model address the class imbalance issue. We gain more improvements by adding the dice loss as it helps the model maximize the region overlap between the predictions and the ground truth.  

% \input{tables/loss_ablation}

% \subsection{3-D Reconstructions}

% \JKR{It would a great addition if you could do a full 3D reconstruction over the whole 512x256 field of view}

\end{document}